# Exploiting Language Relatedness in Machine Translation Through Domain Adaptation Techniques

Amit Kumar, Rupjyoti Baruah, Ajay Pratap, *Member, IEEE*, Mayank Swarnkar, and Anil Kumar Singh

*Abstract*—One of the significant challenges of Machine Translation (MT) is the scarcity of large amounts of data, mainly parallel sentence aligned corpora. If the evaluation is as rigorous as resource-rich languages, both Neural Machine Translation (NMT) and Statistical Machine Translation (SMT) can produce good results with such large amounts of data. However, it is challenging to improve the quality of MT output for low resource languages, especially in NMT and SMT. In order to tackle the challenges faced by MT, we present a novel approach of using a scaled similarity score of sentences, especially for related languages based on a 5-gram KenLM language model with Kneser-ney smoothing technique for filtering in-domain data from out-of-domain corpora that boost the translation quality of MT. Furthermore, we employ other domain adaptation techniques such as multi-domain, fine-tuning and iterative back-translation approach to compare our novel approach on the Hindi-Nepali language pair for NMT and SMT. Our approach succeeds in increasing ~2 BLEU point on multi-domain approach, ~3 BLEU point on fine-tuning for NMT and ~2 BLEU point on iterative back-translation approach.

## I. INTRODUCTION

Machine Translation *(MT)*, by its nature, is a challenging problem as it has to deal with various forms of linguistics irregularities in many languages simultaneously [1]. To deal with irregularities in languages, MT requires a massive corpus. Corpus is a language resource that has a collection of large and unstructured text stored on the computer. Recent research on Neural Machine Translation *(NMT)* suggests that it achieves parity with human translation on some High Resource Languages *(HRLs)*, especially true for pairs like English-German and English-French [2]. NMT models have shown awe-inspiring performance for HRLs. However, there are some languages for which sufficient resources are not available, called Low Resource Languages *(LRLs)*. Nowadays, for LRLs, various techniques have been developed to perform well in a low-resource scenario. But there are some issues with approaches designed for LRLs. One reason for this is that evaluating such languages is more rigorous than evaluating resource-rich languages due to the low quantity and quality of data. Another problem is that the results are still not practically anywhere near as usable as for HRLs. To handle such problems, using Transfer Learning *(TL)* with similar languages is one of the promising solutions in the direction of MT [3]. The process of adapting the features from HRLs to LRLs, known as TL [4]. Domain Adaptation *(DA)* is a type of TL that applying the model trained on source distribution in the context of different, but related target domains [5]. Sometimes, the word "Domain" is used for a language among a set of similar languages, and DA used for adapting the features of HRLs to LRLs.

Even so, the techniques developed for NMT in LRLs are worthwhile to explore further as they do show improvement. A fortunate fact in this regard is that numerous pairs of such languages are more or less closely related in linguistic terms such as Marathi-Hindi and Nepali-Hindi [6], [7]. In addition, it is often the case that most such pairs have at least one closely related language in terms of lexicon, morphological features, syntax and semantics. Using techniques like DA, we can leverage this similarity of close languages for better low resource MT. In this paper, we do precisely this for the Nepali-Hindi pairs in both translation directions. Hindi *(HI)* and Nepali *(NE)* belong to the common origin of languages: the Indo-Aryan branch of the Indo-European family. Hindi written in the Devanagari script [8]. Nepali, also written in Devanagari script, is spoken mainly in Nepal and some parts of Bhutan and India. Since people of these two languages have been in contact for millennia, there is an exchange and borrowing of cognates [9], making the two languages even more similar to each other. Apart from the similarities in the vocabularies, these languages have the same word order and orthographic features. The similarity of lexical and phonological features between two languages demonstrates with the help of overlapping of words (Table VIII). As Hindi and Nepali are Indic languages, there is a need to bridge the communication gap between these two. Developing translation tools acts as a destroyer of communication barriers between human relations.

DA problems have been well studied in MT for many language pairs. Most existing methods focus on using DA at model levels and sentence-level approaches given in [10], [11], [12] and [13]. However, there is no specific approach used for similar or related languages. Therefore, a need to propose an approach specially designed for

A. Kumar, R. Baruah, A. Pratap, M. Swarnkar and A. K. Singh are with the Department of Computer Science and Engineering, Indian Institute of Technology (Banaras Hindu University) Varanasi 221005 India. E-mail: {amitkumar.rs.cse17, rupjyotibaruah.rs.cse18, ajay.cse, mayank.cse, aksingh.cse}@iitbhu.ac.in.

similar language can bring a drastic improvement in MT.

In order to solve the DA problem, we proposed scaled similarity score-based sentence selection approach for similar language on a sentence level. In sentence selection, the hypothesis is that a language model trained on the corpus encoded in WX-notation [14] helps better learn the similarity between out-of-domain and in-domain sentences. Compared to other data selection approaches, we compute the Scaled Similarity Score *(SSS)* of the sentence based on training 5-gram KenLM language model with Kneser-Ney smoothing technique [15].

The contributions of the paper are summarized as:

1) As per best of our knowledge, we are the first to use the *SSS* of the sentence instead of cross-entropy as a data selection approach for filtering the in-domain sentences from out-of-domain corpora.
2) Using the WX-transliteration scheme in training the language model helps better learn the representation of sentences for similar languages.
3) Using scaled similarity score-based data selection approach with Iterative Back-Translation (IBT) outperform the current state-of-the-art models.
4) Solve the DA problem efficiently by designing algorithm for proposed models that complete in polynomial time.

The rest of the paper is organized as follows. Section II discusses the closely related works. Section III describes the proposed solutions with computational complexity. Performance study with result analysis, corpus statistics and the experimental setup required to conduct the experiments are reported in Section IV. At last, we conclude the papers in Section V with future aspects of our work.

Table I: Comparison of related work

| Paper | SMT | NMT | SSS | Language pair |
|---|---|---|---|---|
| [10] | ✗ | ✗ | ✗ | EN-FR |
| [11] | ✓ | ✗ | ✗ | ZH-EN |
| [16] | ✓ | ✗ | ✗ | EN-DE, FR, RU, ES |
| [17] | ✓ | ✗ | ✗ | EN-ES |
| [18] | ✓ | ✗ | ✗ | DE-EN, AR-EN |
| [19] | ✓ | ✗ | ✗ | AR-EN, EN-FR |
| [20] | ✓ | ✗ | ✗ | ZH-EN |
| [21] | ✓ | ✗ | ✗ | EN-DE, EN-TR |
| [22] | ✗ | ✓ | ✗ | EN-FR, EN-DE |
| [23] | ✗ | ✓ | ✗ | EN-JA, DE-EN |
| [13] | ✗ | ✓ | ✗ | EN-DE, EN-IT |
| [24] | ✗ | ✓ | ✗ | DE-EN |
| [25] | ✗ | ✓ | ✗ | ZH-EN, EN-FR, EN-DE |
| SSS | ✓ | ✓ | ✓ | HI-NE |

Note- EN: English, FR: French, ZH: Chinese, DE: German, RU: Russian, ES: Spanish, AR: Arabic, TR: Turkish, JA: Japanese, IT: Italian

## II. RELATED WORKS

There are many works done in DA applied to Statistical Machine Translation *(SMT)* and NMT models [26]. These works are categorized into data-centric and model-centric. Data-centric techniques include selecting training data from out-of-domain parallel data based on a Language Model *(LM)* or generating pseudo parallel data [10], [11], [16]–[18]. In contrast, the model-centric approach focuses on the injection of in-domain and out-of-domain models at the level of either a model or instance [19]–[21]. In the following we review the closely related works with data-centric and model-centric approaches.

### A. Data-centric approach

In [10], authors proposed a data selection approach which computes the language model cross-entropy difference for each sentence in a monolingual corpus. Cuong et.al [17] derived an invitation model based on iterative weighted invitations using Expectation Maximization *(EM)* algorithm, offers a data selection approach for MT on a large parallel corpus consisting of a mix of a rather diverse set of domains. In [18], authors applied the EM-based mixture modeling and data selection techniques using two joint models, namely the operation sequence model — an n-gram based translation and reordering model, and the neural network joint model — a continuous space translation model, to carry out DA for MT. However, these approaches does not give satisfactory result if language model trained on extremely low resource data.

In [22], authors talk about sentence-level DA for NMT. They exploit the NMT's internal sentence embedding and used the sentence embedding similarity to select out-of-domain sentences that are close to the in-domain corpus. They also proposed dynamic training method and multi-domain sentence weighting method to balance the domain distributions of training data and match the domain distributions of training and testing data. However, these approaches give good contributions but they does not talk about leveraging the orthographic features of languages in training the model.

Dou et.al [13] proposed dynamic data selection and weighting approach for iterative back-translation. In [24], authors proposed Classifier Augmented Filtered Iterative Back-Translation (CFIBT) for the domain adaptation task. They trained two convolutional neural network based binary classifiers, one in source and the other in the target language on the combination of in-domain and out-of-domain corpora. They used IBT for synthetic parallel corpora generation and classifier-based filtering to remove the pair of sentences where the synthetic sentence in the pair does not belong to the domain. These approaches focused on extending the selection approaches to IBT but they does not talk about leveraging the orthographic features and relatedness between languages in data selection approaches.

### B. Model-centric approach

In [19], authors proposed the fill-up adaptation method on a popular open source SMT platform, tested it on a speech translation task. Zhou et.al [20] described sentence-weight-based adaptation approach depending on the similarity between sentences in the training set and the target domain text. In [21], authors proposed the FDA5, a parameterization, optimization, and implementation framework for feature decay algorithms (FDA), a class of instance selection algorithms that employ feature decay.

Sato et.al [23] proposed vocabulary adaptation, a simple method for effective fine-tuning that adapts embedding layers in a given pretrained NMT model to the target domain. Prior to fine-tuning, their method replaces the embedding layers of the NMT model by projecting general word embeddings induced from monolingual data in a target domain onto a source-domain embedding space. In [25], authors first prune the model and only keep the important neurons or parameters, making them responsible for both general domain and in-domain translation. They further trained the pruned model supervised by the original whole model with knowledge distillation. At last they expand the model to the original size and fine-tune the added parameters for the in-domain translation.

However, all the models discussed above does not consider the orthographic features of the text during language model training.

### C. Shortcomings of existing methods

Most of the existing methods required extra monolingual data of specific domains to train the language model. Language model needs training data in sufficient amounts for better learning, and data is the major challenge faced by any low resource MT models. Contrary to the existing superior methods, our approach, despite based on a statistical language model, does not need extensive data to train the model. Our approach leverage the orthographical features of text by transliterating them into WX-notation.

Table I lists the comparison of our proposed approach with other related works. In comparison to existing approaches, our model is the first to use the SSS of the sentence based on the Kneser-Ney smoothing technique for data selection. Furthermore, our model trained on a 5-gram KenLM and work as a filter to separate in-domain from out-of-domain data.

Table II: Naming conventions

| Conventions | Domain |
|---|---|
| In-domain | WMT2019 |
| AGRI | Agriculture |
| ENT | Entertainment |
| BIB | Bible |
| $MD_1$ | WMT+Agriculture |
| $MD_2$ | WMT+Entertainment |
| $MD_3$ | WMT+Bible |
| $MD_4$ | Agriculture+Entertainment |
| $MD_5$ | Agriculture+Bible |
| $MD_6$ | Entertainment+Bible |
| $MD_7$ | WMT+Agriculture+Entertainment |
| $MD_8$ | WMT+Agriculture+Bible |
| $MD_9$ | WMT+Entertainment+Bible |
| $MD_{10}$ | Agriculture+Entertainment+Bible |
| $MD_{11}$ | WMT+Agriculture+Entertainment+Bible |

## III. Proposed Model

In this section, we discuss our proposed model in order to solve the DA problem of improving the MT quality by leveraging the similarity between the out-of-domain and in-domain sentences. Table II lists the domains used in our experiments. Our model focuses on using SSS data selection approach for DA. The architecture of proposed model is shown in Fig. 1. Our model takes corpora of multiple domains (agriculture, entertainment, and bible) as input for filtering in-domain sentences using SSS. SSS is computed by training KenLM on WX-transliterated in-domain corpus using (1). SSS is applied on the source side of each sentences in multiple domains and arrange the sentences in order of score obtained. Then model selects the sentences satisfying the threshold condition and stored the selected sentences as a set of filtered data. It concatenates the filtered data to in-domain corpus and stored the concatenated data as parallel corpus (source and target) for further training of MT model. Finally, MT gives the predicted translated sentences. In addition to this, we also design the algorithm and propose the time complexity of multi-domain, fine-tuning, and IBT to compare the performance of approaches which is described in the following section.

### A. Multi-domain approach

In this section, we focus only on a corpus-based approach. Algorithm 1 describes how the models trained on different combinations of corpora. We employed training data of mixed domains and validation data of in-domain as input to the Multi-Domain Algorithm *(MDA)*. MDA at final results the list of trained MT models ($MT_{s \to t,k}$, $MT_{t \to s,k}$) for all $k \in$ *Training Data* in both directions Source ($s$)→Target ($t$) and Target ($t$)→Source ($s$). Firstly, MDA calls each training data and performs training on model $MT_{s \to t}$ for $s \to t$ (lines 1-2). After training $MT_{s \to t}$, it trains the model $MT_{t \to s}$ in the direction of $t \to s$ (line 3). Both the models continue training until it converges. Then MDA add the both models in list (line 4). After calling each training data, MDA returns the list($MT_{s \to t,k}$, $MT_{t \to s,k}$) for all $k \in$ *Training Data*.

---

**Algorithm 1:** Multi-domain Approach

**Input:** *Training_Data* ∈ list(In-domain, AGRI, ENT, BIB, $MD_1$, $MD_2$, $MD_3$, $MD_4$, $MD_5$, $MD_6$, $MD_7$, $MD_8$, $MD_9$, $MD_{10}$, $MD_{11}$), *Valid_Data* ∈ In-domain, l = list()

**Output:** list($MT_{s \to t,k}$, $MT_{t \to s,k}$), ∀k ∈Training_Data

1 **for** ∀k ∈ *Training_Data* **do**
2    $MT_{s \to t} \leftarrow$ Train MT using *Training_Data*, *Valid_Data* for s → t;
3    $MT_{t \to s} \leftarrow$ Train MT using *Training_Data*, *Valid_Data* for t → s;
4    l.append($MT_{s \to t}$, $MT_{t \to s}$);

---

### B. Fine-tuning multi-domain on in-domain

Algorithm 2 describes the training process of Fine-Tuning Approach *(FTA)*. We employed parallel training data of Out-of-domain ($OD_p$)and In-domain ($ID_p$), and validation data of in-domain as input to the FTA. Finally, FTA results the list of trained NMT models ($NMT_{s \to t,b}$, $NMT_{t \rightarrow s,b}$) $\forall b \in OD_p$ in both directions $s \rightarrow t$ and $t \rightarrow s$. Firstly, FTA calls each element of the $OD_p$ and performs training on model $NMT_{s \rightarrow t}$ for direction $s \rightarrow t$ (lines 1-2). After training $NMT_{s \rightarrow t}$, it trains the model $NMT_{t \rightarrow s}$ in the direction of $t \rightarrow s$ (line 3). Both the models continue training until it converges. Then it trains the $NMT_{s \rightarrow t}$ on $ID_p$ for direction $s \rightarrow t$ using the model weights gain from above model $NMT_{s \rightarrow t,b}$ (line 4). Similarly, it trains $NMT_{t \rightarrow s}$ on $ID_p$ for direction $t \rightarrow s$ using the model weights gain from above model $NMT_{t \rightarrow s,b}$ (line 5). Then FTA adds both models in list (line 6). After calling each element of $OD_p$, FTA returns the list($NMT_{s \rightarrow t,b}$, $NMT_{t \rightarrow s,b}$) $\forall b \in OD_p$.

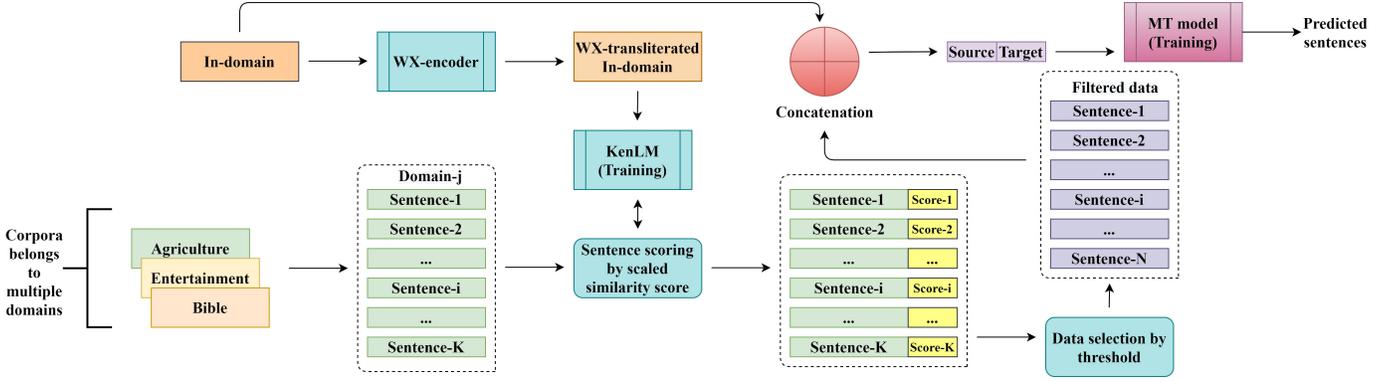

Figure 1: Proposed architecture.

DA for Scaled Similarity Score-based Selection Approach (DASSA). DASSA at final results the trained MT models ($MT_{s \rightarrow t}$, $MT_{t \rightarrow s}$) in both directions $s \rightarrow t$ and $t \rightarrow s$. Firstly, DASSA trains KenLM on $s$ and $t$ (lines 1-2). After training KenLM, DASSA calls each sentence of $MD_{10}$ and compute $SSS$ for both $s$ and $t$ (lines 3-5). Then it checks whether $SSS$ for both $s$ and $t$ is greater than or equal to some threshold and if $SSS$ satisfy the threshold condition, DASSA add the sentence to filtered in-domain corpora ($ID'_p$ or $ID''_p$) (lines 6-9). DASSA adds the filtered in-domain corpora to $ID_p$ and trains the model $MT_{s \rightarrow t}$ and $MT_{t \rightarrow s}$ (lines 10-13). Finally, it returns the $MT_{s \rightarrow t}$ and $MT_{t \rightarrow s}$.

---

**Algorithm 2:** Fine-tuning Approach

**Input**: $OD_p \in$ list(AGRI, ENT, BIB, $MD_4$, $MD_5$, $MD_6$, $MD_{10}$), $ID_p \in$ In-domain, $Valid\_Data \in$ In-domain, l = list()
**Output**: list($NMT_{s \rightarrow t,b}$, $NMT_{t \rightarrow s,b}$), $\forall b \in OD_p$

1 **for** $\forall b \in OD_p$ **do**
2    $NMT_{s \rightarrow t} \leftarrow$ Train NMT using $OD_p$ as training_data, $Valid\_Data$ for $s \rightarrow t$;
3    $NMT_{t \rightarrow s} \leftarrow$ Train NMT using $OD_p$ as training_data, $Valid\_Data$ for $t \rightarrow s$;
4    $NMT_{s \rightarrow t} \leftarrow$ Train NMT on $ID_p$ using the model weights from $NMT_{s \rightarrow t,b}$ for $s \rightarrow t$;
5    $NMT_{t \rightarrow s} \leftarrow$ Train NMT on $ID_p$ using the model weights from $NMT_{t \rightarrow s,b}$ for $t \rightarrow s$;
6    l.append($NMT_{s \rightarrow t}$, $NMT_{t \rightarrow s}$);

---

### C. Domain filtering approach

We are using a SSS to obtain the score of $i^{th}$ sentence ($S_i$) in the corpus ($C$) which is computed as follows:

$$SSS(S_i) = \frac{L_i - min(L_S)}{max(L_S) - min(L_S)} \qquad (1)$$

where, $L_i$ is the score computed by KenLM for $S_i$, $min(L_S)$ and $max(L_S)$ are the minimum and maximum score computed by KenLM for sentence in $C$, respectively.

Algorithm 3 uses a SSS to separate in-domain data from out-of-domain corpora by training KenLM on WX-transliterated source and target-side of the sentences of in-domain corpus. We employed $MD_{10}$ and $ID_p$ as training data, and validation set of in-domain as input to the

---

**Algorithm 3:** DA for SSS Selection Approach

**Input**: $MD_{10}$, $ID_p \in$ In-domain, $Valid\_Data \in$ In-domain
**Output**: $MT_{s \rightarrow t}$, $MT_{t \rightarrow s}$

1 Train KenLM($s \in ID_p$) for $s \rightarrow t$ ;
2 Train KenLM($t \in ID_p$) for $t \rightarrow s$ ;
3 **for** $\forall j \in MD_{10}$ **do**
4    S(j) $\leftarrow SSS(j)$ for $s \rightarrow t$;
5    S'(j) $\leftarrow SSS(j)$ for $t \rightarrow s$;
6    **if** $S(j) \geq threshold1$ **then**
7      $ID_p' \leftarrow$ j;
8    **if** $S'(j) \geq threshold2$ **then**
9      $ID_p'' \leftarrow$ j;
10 $ID_p' \leftarrow ID_p + ID_p'$;
11 $MT_{s \rightarrow t} \leftarrow$ Train MT using $ID_p'$ for $s \rightarrow t$;
12 $ID_p'' \leftarrow ID_p + ID_p''$;
13 $MT_{t \rightarrow s} \leftarrow$ Train MT using $ID_p''$ for $t \rightarrow s$;

---

### D. IBT with SSS selection approach

We use the SSS method in IBT approach to filter the monolingual corpus closed to the in-domain data. Algorithm 4 describes Scaled Similarity Score-based Iterative Back-translation Approach (SSSIBA). We employed source and target monolingual corpora ($M_s$ and $M_t$), and $ID_p$ corpora as training data, and validation data of in-domain as input to SSSIBA. Finally, SSSIBA results the trained NMT models ($NMT_{s \rightarrow t}$, $NMT_{t \rightarrow s}$) in both directions $s \rightarrow t$ and $t \rightarrow s$. Firstly, SSSIBA trains KenLM on both source and target side of $ID_p$ (lines 1-2). Then it trains $NMT_{s \rightarrow t}$ and $NMT_{t \rightarrow s}$ using in-domain parallel corpora $ID_p$ (lines 4-5). After training, SSSIBA translates $M_t$ and $M_s$ using $NMT_{t \to s}$ and $NMT_{s \to t}$ respectively (lines 4-5). Then SSSIBA performs filtering on translated monolingual data ($M'_s$ and $M'_t$) and it creates synthetic parallel corpus ($SN'_s$ and $SN'_t$) (lines 6-15). Then concatenating the synthetic parallel corpus with $ID_p$ and creates the $ID'_p$ and $ID''_p$ (lines 16-17). Finally, SSSIBA trains the $NMT_{s \to t}$ and $NMT_{t \to s}$ using $ID'_p$ and $ID''_p$ respectively (lines 18-19). This entire process repeats until convergence.

---

**Algorithm 4:** IBT with SSS Selection Approach

**Input:** $M_t \in$ monolingual(t), $M_s \in$ monolingual(s), $ID_p \in$ list(In-domain), $Valid\_Data \in$ In-domain

**Output:** list($NMT_{s \to t}$, $NMT_{t \to s}$)

1   Train KenLM(s $\in ID_p$) for s $\to$ t ;
2   Train KenLM(t $\in ID_p$) for t $\to$ s ;
3   **while** $NMT_{s \to t}$ and $NMT_{t \to s}$ not converged **do**
4      $M'_s \leftarrow$ Translate $M_t$ using $NMT_{t \to s}$;
5      $M'_t \leftarrow$ Translate $M_s$ using $NMT_{s \to t}$;
6      **for** $\forall j \in M'_s$ **do**
7          S(j) $\leftarrow SSS(j)$ for s $\to$ t;
8          **if** $S(j) \geq threshold1$ **then**
9              $F_s \leftarrow$ j;
10      **for** $\forall j \in M'_t$ **do**
11          S(j) $\leftarrow SSS(j)$ for t $\to$ s;
12          **if** $S(j) \geq threshold2$ **then**
13              $F_t \leftarrow$ j;
14      $SN'_s \leftarrow (F_s, M_t)$;
15      $SN'_t \leftarrow (M_s, F_t)$;
16      $ID'_p \leftarrow ID_p + SN'_s$;
17      $ID''_p \leftarrow ID_p + SN'_t$;
18      $NMT_{s \to t} \leftarrow$ Train NMT using $ID'_p$ for s $\to$ t;
19      $NMT_{t \to s} \leftarrow$ Train NMT using $ID''_p$ for t $\to$ s;

---

### E. Analysis of proposed algorithm

In this section, we discuss convergence and time complexity of proposed model. The time complexity of the proposed model depends on the complexity of MT modules it calls. The time complexity of KenLM is $O(\log \log K)$ where $K$ is number of key-value pairs in sorted array [15]. The time complexity of Transformer-NMT is $O(W^2 D + WD^2)$ where $W$ is number of word-vectors and $D$ is the dimension of each word-vector [27]. The time complexity of SMT depends on complexity of KenLM, Moses decoder, and reordering model. The time complexity of decoding with histogram pruning is $O(TL^2)$ where $T$ is the maximum stack size of hypothesis and $L$ is the length of sentences [28]. The time complexity of reordering model is $O(TL)$ [28]. Thus, the time complexity of SMT is $O(T^2 L^3 \log \log K)$.

Time complexity of each algorithm is described in the following section:

*1) MDA:* The time complexity of MDA using SMT and NMT are $O(GT^2 L^3 \log \log K)$ and $O(GW^2 D + GWD^2)$ respectively where $G$ is the number of domains used for training (lines 1-3 from Algorithm 1).

*2) FTA:* The time complexity of FTA is $O(HW^2 D + HWD^2)$ where $H$ is the number of out-of-domains used for training (lines 1-5 from Algorithm 2).

*3) DASSA:* The time complexity of Algorithm 3 from lines 1-2 is $O(\log \log K)$, lines 3-9 is $O(N)$, and lines 10-13 are $O(T^2 L^3 \log \log K)$ for SMT and $O(W^2 D + WD^2)$ for NMT. Thus, the time complexity of DASSA using SMT and NMT are $O(N + \log \log K + T^2 L^3 \log \log K))$ and $O(N + \log \log K + W^2 D + WD^2)$ respectively where $N$ is the number of sentences in $MD_{10}$ domain.

*4) SSSIBA:* The time complexity of Algorithm 4 from lines 1-2 is $O(\log \log K)$, lines 4-5 and 18-19 is $O(W^2 D + WD^2)$, and lines 6-13 is $O(M)$. Thus, the time complexity of SSSIBA is $O(\log \log K + W^2 ID + WID^2 + IM)$ where $M$ is the number of sentences in monolingual corpus and $I$ is the number of iterations required to converge the model.

*5) Convergence of model:* We consider MT models to be converged when there is no improvement in the models upto continuous 10 iterations.

## IV. Performance Study

In this section, we discuss the corpus statistics and experimental setup required to execute the experiments.

### A. Corpus description

The parallel corpora for training, validation and testing of in-domain Hindi-Nepali data were gathered from a similar language shared task organized in the Workshop on Machine Translation (WMT 2019) [7]. In addition, the Agriculture and Entertainment domains came from TDIL [29], and the Bible domain and monolingual data were collected from OPUS [30]. Table III contains details on the statistics of corpora belongs to various domains.

Table III: Corpus description

| Domain | Training Sentences |
|---|---|
| In-domain | 65505 |
| AGRI | 14000 |
| ENT | 27000 |
| BIB | 30486 |
| $MD_1$ | 79505 |
| $MD_2$ | 92505 |
| $MD_3$ | 95991 |
| $MD_4$ | 41000 |
| $MD_5$ | 44486 |
| $MD_6$ | 57486 |
| $MD_7$ | 106505 |
| $MD_8$ | 109991 |
| $MD_9$ | 122991 |
| $MD_{10}$ | 71486 |
| $MD_{11}$ | 136991 |
| Monolingual (HI) | 90000 |
| Monolingual (NE) | 90000 |

### B. Experimental setup

In this section, we discuss different frameworks and parameters required to train the model.

*1) NMT:* For experiments, we use *Fairseq* [31]; a sequence modelling toolkit to train the model and we execute experiments on an *Nvidia V100 GPU*. We used a *transformer* architecture to train the NMT model [27]. We used the *SentencePiece* library for pre-processing [32], and model learns joint vocabulary. The *decoder* and *encoder layers* have been set to 5. Embeddings are shared between the encoder, decoder, and output, implying that our model uses a shared dictionary and embedding space, which should be useful for similar languages and DA. The *encoder* and *decoder embedding dimensions* in the feed-forward network are set to 2048. The *embedding dimensions* of the encoder and decoder are 512. The *attention heads* for the decoder and encoder are set to 2. The models are regularized with *dropout, label smoothing* and *weight decay*, with the corresponding hyper-parameters being set to 0.4, 0.2 and 0.0001, respectively. Models were optimized with *Adam* using $\beta 1 = 0.9$ and $\beta 2 = 0.98$ and have *patience* value 10.

*2) SMT:* We use *Moses*, open-source toolkit to train SMT [32]. For creating phase/words alignment from a *Hindi-Nepali* parallel corpus, we use *GIZA++* [34]. A 5-gram KenLM language model is used for training [15]. The parameters are tuned on the validation set using *MERT* and tested with a test set [35].

*3) Filtering Model:* We trained *5-gram* language model on in-domain corpus using KenLM. We keep the threshold value to 0.8 for filtering the in-domain data from out-of-domain data.

## C. Result analysis

To evaluate the performance on different domains, we used *BLEU* [36], [37], *chrF2* [38] and *TER* [39], as shown in Tables V and VII. We demonstrate our approach on different setup using SMT and NMT. BLEU score is a standard metric accepted by *NLP* researchers to obtain the accuracy of predicted translated outputs compared to the human-translated reference sentences (gold labels) [36]. It is observed that the higher the value of the BLEU score, better the output of translations. The formula of the BLEU score is as follows [36]:

$$BLEU = min\left(1, \frac{output\_length}{reference\_length}\right)\left(\prod_{i=1}^{4} precision_i\right) \quad (2)$$

where *output_length* and *reference_length* are the length of predicted sentences and reference sentences respectively.

Since $MD_{10}$ consist of three out-of-domain corpus: agriculture, entertainment, and bible. We apply our filter approach on $MD_{10}$ to compare the result with other existing approaches.

*1) Applying WX notation in language model:* We compute the perplexity of KenLM language model trained on source and target side of in-domain corpus using WX-transliteration and without WX-transliteration. Perplexity scores are listed in Table IV. We find that perplexity of language model trained on training data transliterated to WX-notation is less than the language model trained on training data without transliteration. Language model having low perplexity is always better. It means WX-transliterated language model performs better than language model trained on without WX-transliterated cor-pus. Therefore, we can say that our data selection approach based on WX-transliterated trained language model performs better than the existing approaches.

Table IV: Perplexity of KenLM

| Corpus | Perplexity |
|---|---|
| Source side (HI) | 477.324 |
| Target side (NE) | 550.333 |
| Source side (HI)[*] | 8.162 |
| Target side (NE)[*] | 6.756 |

[*] Data is transliterated into WX-notation

*2) Effect of using SSS method on SMT and NMT :* Table V contains the results on applying *SSS* for domain filtering and comparing it with SMT and NMT, respectively, by training the model on the number of corpora, consist of merging the data of multiple domains. Domain filtering is applied on purely out-of-domain corpora of the $MD_{10}$ domain (corpora belong to Agriculture, Entertainment, and Bible domains). Our approach beats the MDA performance by ∼2.2 around on SMT and on an average ∼2 on NMT. On comparing to Moore et.al [10] data selection approach on MD10, we get an improvement of ∼1 on SMT and ∼2 on NMT. The reason for such improvement is SSS data selection approach depends on WX-transliterated trained language model. WX-transliteration decreases the language distance between out-of-domain and in-domain corpus (Table IV). Another reason is scaling the score with SSS reduces the large gap between minimum and maximum score of sentences (score lies in the range of 0 to 1). So it favors the selecting more sentences close to the in-domain corpus.

*3) Effect of using fine-tuning on related languages:* Table VI contains the result of fine-tuning the models trained using the corpora consist of merging of out-of-domain corpora on in-domain data. Fine-tuning improves translations on NMT by a small margin when compared to models without fine-tuning. The reason for such output is that out-of-domain data is made up of multiple domains. Data from multiple domains limit the model's ability to perform better because of large divergence in representation of data. Our data selection approach reduces the data complexity from out-of-domain corpora and gives improvement of ∼3 on fine-tuning and ∼2 on applying Moore et.al [10] data selection approach on $MD_{10}$. Reason for such good improvement is same as we explain in previous section. SSS scaling and using WX-transliteration favors the model in selection more sentences close to the in-domain corpus.

*4) Effect of using back-translation:* We integrate the *SSS* data selection approach with IBT on NMT. We compare it with one iteration Back-Translation (BT) and dynamic data selection approach used by Dou et.al [13] with TF-IDF on Nepali-Hindi language pair. Table VII lists the comparison of our integrated IBT approach with BT and dynamic data selection. Our integrated IBT approach outperforms the BT with ∼2 and [13] with ∼1

Table V: Data selection results

| Models | NE→HI | | | HI→NE | | |
|---|---|---|---|---|---|---|
| | BLEU | chrF2 | TER | BLEU | chrF2 | TER |
| MDA (SMT) | 43.91 | 0.53 | 0.601 | 47.11 | 0.57 | 0.562 |
| [10] (SMT) | 45.88 | 0.54 | 0.579 | 48.91 | 0.57 | 0.545 |
| DASSA (SMT) | 47.01 | 0.54 | 0.578 | 49.31 | 0.58 | 0.540 |
| MDA (NMT) | 30.51±0.04 | 0.49±0.00 | 0.645±0.000 | 32.57±0.64 | 0.52±0.00 | 0.624±0.000 |
| [10] (NMT) | 30.80±0.01 | 0.49±0.00 | 0.640±0.00 | 31.66±0.13 | 0.51±0.00 | 0.627±0.00 |
| DASSA (NMT) | 32.13±0.55 | 0.51±0.00 | 0.642±0.00 | 35.41±0.23 | 0.53±0.00 | 0.620±0.000 |

Note: NMT models are trained on BPE size of 4000, 5000, and 6000.

Table VI: Results on Fine-tuning

| Domain | NE→HI | | | HI→NE | | |
|---|---|---|---|---|---|---|
| | BLEU | chrF2 | TER | BLEU | chrF2 | TER |
| AGRI[*] | 15.69±0.02 | 0.34±0.00 | 0.791±0.000 | 16.60±0.21 | 0.35±0.00 | 0.789±0.000 |
| ENT[*] | 19.29±0.05 | 0.35±0.00 | 0.803±0.000 | 21.81±0.04 | 0.36±0.00 | 0.797±0.000 |
| BIB[*] | 7.75±0.00 | 0.28±0.00 | 0.796±0.000 | 10.50±0.02 | 0.29±0.00 | 0.790±0.000 |
| MD4[*] | 17.48±0.31 | 0.36±0.00 | 0.770±0.000 | 17.56±0.44 | 0.36±0.00 | 0.756±0.000 |
| MD5[*] | 19.21±0.03 | 0.37±0.00 | 0.759±0.000 | 21.70±0.01 | 0.37±0.00 | 0.769±0.000 |
| MD6[*] | 19.93±0.34 | 0.38±0.00 | 0.752±0.000 | 20.70±0.23 | 0.39±0.00 | 0.740±0.000 |
| MD10[*] | 16.91±0.04 | 0.36±0.00 | 0.758±0.000 | 18.32±0.42 | 0.37±0.00 | 0.742±0.000 |
| MD10 [10] | 17.88±0.03 | 0.37±0.00 | 0.756±0.000 | 19.12±0.67 | 0.38±0.00 | 0.739±0.000 |
| MD10[**] | 19.21±0.13 | 0.38±0.00 | 0.747±0.000 | 21.31±0.05 | 0.40±0.00 | 0.731±0.000 |

Note: NMT models are trained on BPE size of 4000, 5000, and 6000.
[*]: Using FTA approach.
[**] Applying SSS approach on MD10 domain and then performing FTA.

Table VII: Back-translation with data selection

| Domain | NE→HI | | | HI→NE | | |
|---|---|---|---|---|---|---|
| | BLEU | chrF2 | TER | BLEU | chrF2 | TER |
| BT[*] | 21.17±0.04 | 0.36±0.00 | 0.699±0.00 | 23.67±0.14 | 0.41±0.00 | 0.645±0.000 |
| Dou et.al [13] | 21.88 | 0.37 | 0.670 | 24.75 | 0.42 | 0.641 |
| SSSIBA | 22.91±0.15 | 0.39±0.00 | 0.589±0.00 | 25.88±0.27 | 0.44±0.00 | 0.561±0.00 |

Note: NMT models are trained on BPE size of 4000, 5000, and 6000.
[*] Back-translation using monolingual data on NMT.

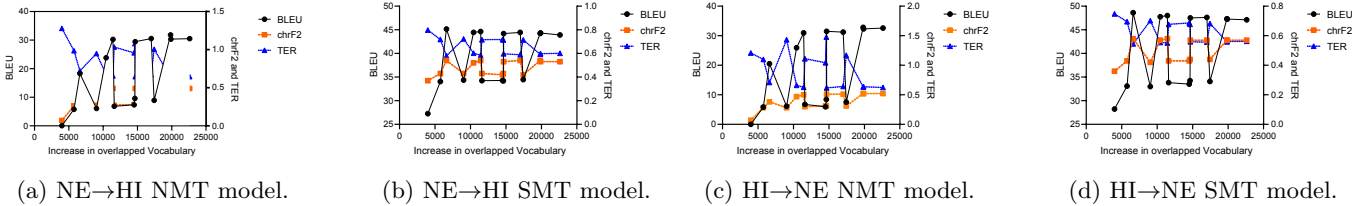

(a) NE→HI NMT model.    (b) NE→HI SMT model.    (c) HI→NE NMT model.    (d) HI→NE SMT model.

Figure 2: Overlapped vocabulary versus different evaluation metrics (BLEU, chrF2, and TER).

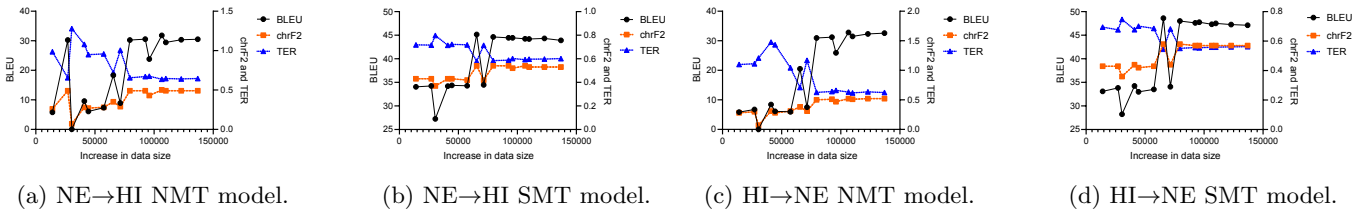

(a) NE→HI NMT model.    (b) NE→HI SMT model.    (c) HI→NE NMT model.    (d) HI→NE SMT model.

Figure 3: Data size versus different evaluation metrics (BLEU, chrF2, and TER).

BLEU point. Improvement in translation after integration of SSS in IBT again shows that scaling of scores using SSS and WX-transliteration boosts the model performance by selecting more sentences close to the in-domain corpus.

*5) Relatedness between languages:* To quantitatively measure the similarity, we used the *char-BLEU* [40],

*chrF2*, and *TER*, in addition to using them for evaluation. These evaluation metrics can be used for similarity if the data is represented in a common encoding format. So, we encode all of our training data into *WX* encoding [14]. Then, we apply all these metrics to the training corpus of all the domains. More score of *char-BLEU* and $chrF2$ means more similarity between languages. On the other hand, the lesser $TER$ score infer better the similarity between languages. We also used overlapping vocabularies to measure the similarity between Hindi and Nepali for each training data. Domains with more overlapped vocabularies mean Nepali and Hindi corpus of that particular domain is more related or similar. It is done to get more insights into the results of the experiments and their relationships with similarity. Table VIII lists the scores of each metric which represents the similarity between language pairs. The purpose is to report and observe the relations between similarity estimates and the evaluation results. A clear example is the case of the bible. The similarity is the least, and so is the performance of MT. Similarly, for the entertainment domain, both the similarity and the performance of MT are high.

Table VIII: Similarity between NE and HI

| Domain | char-BLEU | chrF2 | TER | OW* |
|--------|-----------|-------|-----|-----|
| In-domain | 0.5255 | 0.47 | 0.857 | 6619 |
| AGRI | 0.5088 | 0.44 | 0.977 | 5757 |
| ENT | 0.5092 | 0.44 | 0.965 | 11622 |
| BIB | 0.2374 | 0.20 | 1.191 | 3998 |
| $MD_1$ | 0.5183 | 0.46 | 0.915 | 11433 |
| $MD_2$ | 0.5163 | 0.45 | 0.926 | 17002 |
| $MD_3$ | 0.3281 | 0.28 | 1.107 | 10431 |
| $MD_4$ | 0.5091 | 0.44 | 0.969 | 14632 |
| $MD_5$ | 0.3119 | 0.26 | 1.140 | 9040 |
| $MD_6$ | 0.3443 | 0.28 | 1.108 | 14516 |
| $MD_7$ | 0.5144 | 0.45 | 0.939 | 19833 |
| $MD_9$ | 0.3843 | 0.32 | 1.064 | 19877 |
| $MD_8$ | 0.3661 | 0.30 | 1.083 | 14672 |
| $MD_{10}$ | 0.3768 | 0.31 | 1.087 | 17440 |
| $MD_{11}$ | 0.4053 | 0.34 | 1.052 | 22620 |

* Number of overlapping words.

We also compare the effect of overlapping vocabularies on NMT and SMT with the help of a line diagram as shown in Fig. 2. We can see the overlapping vocabulary between Hindi and Nepali training sentences. From Fig. 2a, 2b, 2c and 2d, we find that a large degree of overlapping in vocabularies is beneficial to both NMT and SMT except at some domains where data is not close to in-domain corpus. For example, the "$MD_{11}$" domain contains 22620 overlapping words, and it has improved BLEU points compared to the rest of the domains corpus for NMT. However, using our data selection approach on MD10, we achieve improvement of $\sim$2 BLEU point despite large mismatch between training and testing domain data. We can say that sharing vocabulary has a positive impact on NMT but our approach gives good improvement even when sharing vocabulary is not favorable for MT.

*6) Impact of increasing data size:* We demonstrate the effect of increasing training data size on NMT and SMT for related languages with the help of a line diagram as shown in Fig. 3. After going through Fig. 3a and 3c, we find that increasing training data is beneficial for NMT compared to SMT. Furthermore, all the metrics used to evaluate models show co-relation with each other. From Fig. 3b and 3d, we see that SMT achieves saturation of result on in-domain model already.

## V. Conclusion

We have proposed a data selection approach based on SSS that beats the existing state-of-the-art approaches. We used the transformer architecture to train the NMT model and Moses with the KenLM language model for SMT on several domains and compared the result on the different setup of domain adaptation approaches. Our approach gives an increase of $\sim$2 BLEU score on MDA, $\sim$3 BLEU point on FTA, and $\sim$2 BLEU point on IBT approach. We also find that SMT give better translation than NMT on a limited source of data for similar language pair. One of the reasons for such good performance of SMT is an orthographic similarity between the languages.

In future, we will extend our data selection approach with dynamic perspective and integrate with cross-lingual transfer learning with multiple similar languages.